%% file: main.tex
\documentclass[11pt]{article}

\usepackage[preprint]{acl}
\usepackage{tcolorbox}
\usepackage{times}
\usepackage{latexsym}
\usepackage[T1]{fontenc}
\usepackage[utf8]{inputenc}
\usepackage{microtype}
\usepackage{hyperref}
\usepackage{url}
\usepackage{booktabs}
\usepackage{xspace}
\usepackage{graphicx}
\usepackage{textcomp}
\usepackage{multicol}
\usepackage{multirow}
\usepackage{enumitem}
\usepackage{pifont}
\usepackage{amsmath}
\usepackage{subfigure}
\usepackage{float}
\usepackage{tikz}
\usepackage{subcaption}
\usepackage{lineno}
\usepackage{makecell}
\usepackage{CJKutf8}
\usepackage{inconsolata}
\usepackage{adjustbox}

\usepackage{placeins}

\definecolor{darkblue}{rgb}{0, 0, 0.5}
\hypersetup{colorlinks=true, citecolor=darkblue, linkcolor=darkblue, urlcolor=darkblue}

\definecolor{mycolor}{RGB}{33, 95, 154}
\definecolor{custom_red}{RGB}{228, 54, 54}

\title{Reasoning or Fluency? Dissecting Probabilistic Confidence in \\ Best-of-N Selection }

\author{Hojin Kim \\
  Yonsei University \\
  \texttt{rlaghwls@yonsei.ac.kr} \\\And
  Jaehyung Kim \\
  Yonsei University \\
  \texttt{jaehyungk@yonsei.ac.kr} \\}

\begin{document}
\maketitle

\input{0_abstract}
\input{1_intro}
\input{2_related}
\input{3_methodology}

\input{4_results}
\input{5_newmetric}
\input{6_conclusion}
\input{7_limitations}

\bibliography{custom}
\clearpage
\appendix
\input{8_appendix}

\end{document}

%% file: 0_abstract.tex
\begin{abstract}

Probabilistic confidence metrics are increasingly adopted as proxies for reasoning quality in Best-of-$N$ selection, under the assumption that higher confidence reflects higher reasoning fidelity.
In this work, we challenge this assumption by investigating whether these metrics truly capture inter-step causal dependencies necessary for valid reasoning. 
We introduce three classes of inter-step causality perturbations that systematically disrupt dependencies between reasoning steps while preserving local fluency. 
Surprisingly, across diverse model families and reasoning benchmarks, we find that selection accuracy degrades only marginally under these disruptions.
Even severe interventions, such as applying hard attention masks that directly prevent the model from attending to prior reasoning steps, do not substantially reduce selection performance. 
These findings provide strong evidence that current probabilistic metrics are largely insensitive to logical structure, and primarily capture surface-level fluency or in-distribution priors instead. 
Motivated by this gap, we propose a contrastive causality metric that explicitly isolates inter-step causal dependencies, and demonstrate that it yields more faithful output selection than existing probability-based approaches.\footnote{Code to be released upon acceptance}

\end{abstract}

%% file: 1_intro.tex
\section{Introduction}

In recent years, Large Language Models (LLMs) have demonstrated remarkable increases in reasoning capabilities, largely driven by Chain-of-Thought (CoT) prompting and reinforcement learning \citep{jaech2024openai, guo2025deepseek, yang2025qwen3}. 
These capabilities have been shown to scale effectively with increased inference-time computation \citep{snell2024scaling}.
In particular, \textit{Best-of-N}, a strategy that samples multiple reasoning traces and selects the optimal one, has been widely adopted to boost the performance of LLMs across diverse reasoning benchmarks
\citep{wang2022self,lightman2023let,kang2025scalable}. 

Best-of-\textit{N} approaches can be broadly categorized into methods that use an external reward model and reward-free voting schemes.
However, both methods exhibit distinct shortcomings: external reward models incur additional computational costs \citep{wang2024math}, while voting schemes such as self-consistency \citep{wang2022self} suffer from limited applicability to generative tasks \citep{cobbe2021gsm8k, di2025best}. 
Consequently, recent research has shifted towards utilizing \textit{probabilistic confidence metrics} derived directly from the LLM's output token distribution, such as perplexity, as an efficient selection criterion \citep{kang2025scalable,fu2025deep,leang2025picsar}.

\input{Figures/Figure1}

Despite their growing popularity, the reliability of these probabilistic metrics depends on a critical assumption: that they capture the validity of the underlying reasoning process.
However, recent studies challenge this assumption, showing that final answer correctness often does not align with the quality of intermediate reasoning. 
For example, \citet{paul2024making} reveal that correct answers can stem from unfaithful reasoning traces, and \citet{turpin2023language} demonstrate that CoT explanations are frequently post-hoc rationalizations rather than reflections of the model's internal decision-making process.
If CoT traces are often unfaithful or disconnected from the final answer, this raises a fundamental question: \textit{Do probabilistic confidence metrics actually measure reasoning quality, or do they merely capture surface-level fluency?}

To answer this, we investigate {which aspects of CoT traces are truly captured by probabilistic confidence metrics}. 
Specifically, we design three perturbation scenarios that explicitly disrupt \textbf{inter-step causal dependencies} (\textit{i.e.}, the logical flow between reasoning steps) while preserving local fluency. 
We then analyze how confidence metrics respond to these disruptions:
\begin{itemize}[leftmargin=3.5mm, itemsep=3pt, topsep=3pt, parsep=0pt]  
    \item[$\circ$] \textbf{Attention-level disruptions}: We break inter-step causality within the CoT trace by applying attention masking between steps during metric calculation.
    We further test an extreme case by removing the query context during calculation.
    \item[$\circ$] \textbf{Parameter-level disruptions}: We decouple the metric from evaluator reasoning capacity by computing metrics using a substantially smaller model (\textit{e.g.}, \texttt{Qwen2.5-0.5B} \citep{team2024qwen2}) that struggles to verify complex, multi-step logic.
    \item[$\circ$] \textbf{Data-level disruptions}: We structurally disrupt the reasoning trace by truncating, paraphrasing, or shuffling the text, before metric calculation. 
    This preserves local sentence fluency but destroys global logical coherence. 
\end{itemize}

Our experiments reveal a counterintuitive phenomenon: \textit{state-of-the-art probabilistic confidence metrics are surprisingly insensitive to these disruptions}. 
Across five reasoning benchmarks and three LLM families, disrupting inter-step causal structure with the proposed disruptions results in an average selection accuracy loss of less than 1\%, and in several cases even leads to accuracy improvements (see Figure~\ref{fig:fig1}).
These findings provide strong evidence that inherent probability-based metrics act primarily as proxies for \textit{local fluency} and \textit{prior likelihood}, rather than estimating inter-step causality and reasoning quality.
Motivated by this, we propose a \textbf{contrastive causality metric} to isolate signals attributable to inter-step causal structure. 
The proposed metric not only outperforms existing metrics in Best-of-$N$ selection, but also degrades consistently under causal disruptions, indicating greater sensitivity to reasoning coherence.

%% file: Figures/Figure1.tex
\begin{figure}[!t]
  \centering
  \includegraphics[width=1.0\linewidth]{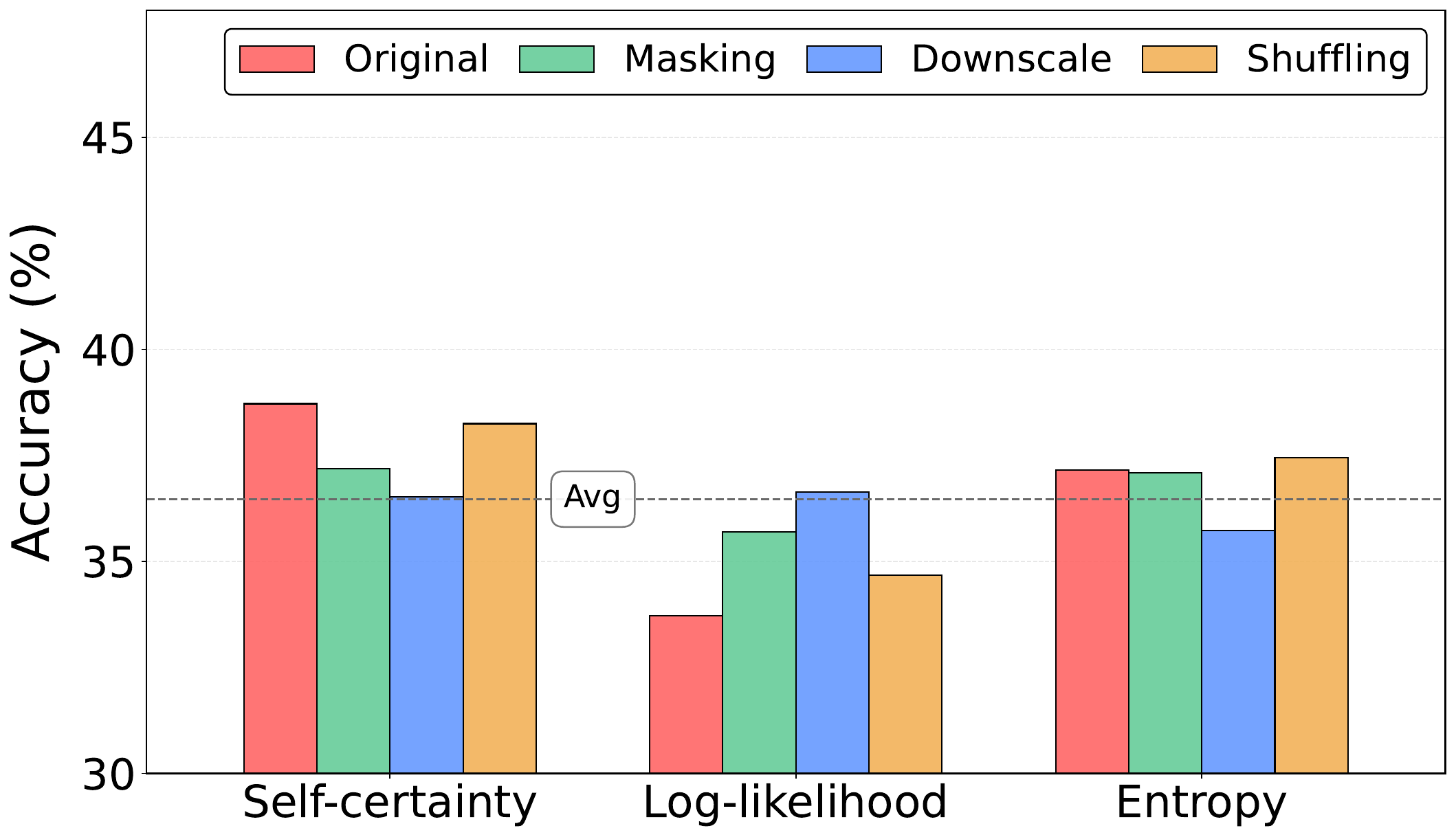}
  \caption{\textbf{Summary of our main results.} Accuracy with Best-of-$N$ selection using three probabilistic confidence metrics (Self-certainty, Log-likelihood, and Entropy) is insensitive to causal disruptions across a number of reasoning benchmarks, implying \textbf{these metrics might not be capturing inter-step causality in CoT traces, but rather local fluency and prior likelihood.}}
  \label{fig:fig1}
\end{figure}

%% file: 2_related.tex
\input{Figures/Figure2}

\section{Related Works}

\paragraph{Best-of-$N$ with probabilistic confidence.}

Existing Best-of-$N$ selection methods using probabilistic confidence metrics fall into two categories: \textit{sentence-level} and \textit{distributional} approaches. 
Methods based on sentence-level confidence evaluate outputs by aggregating the probabilities assigned to sampled tokens. 
While the most straightforward approach simply selects outputs with the highest average log-likelihood (or lowest perplexity), recent research has introduced various refinements.
For example, \citet{leang2025picsar} score candidates using the joint log-likelihood of the reasoning chain and final answer, while \citet{ren2023self} leverages likelihoods assigned by LLMs to different options in a self-evaluation framework. 
Similarly, \citet{fu2025deep} utilize the negative average log-probability of the top-$k$ predicted tokens at each generation step to filter and refine reasoning chains.
An alternative class of methods based on distributional confidence leverages the full output distribution over the entire vocabulary at each step rather than scoring solely based on the sampled tokens.
A natural instantiation of this approach is to use the entropy of the distribution, where a more peaked distribution indicates higher model certainty.
\citet{kang2025scalable} formalize this by aggregating step-wise certainty across the reasoning trace via the KL-divergence between the model’s predicted distribution and a uniform distribution.

\paragraph{Fragility and unfaithfulness of CoTs.} 
Probabilistic confidence metrics are computed over the generated reasoning steps and the final answer.
However, prior work has shown that CoT reasoning steps can be post-hoc or heuristic, and may fail to reflect the model's underlying reasoning process.
For example, \citet{lyu2023faithful} highlight cases where models arrive at the correct final answer without following the preceding steps. 
Similarly, \citet{turpin2023language} demonstrate that models may generate CoT steps that align with surface-level prompt cues and biases, rather than with the model’s true internal decision process.
In addition, \citet{lanham2023measuring} show that the extent to which a model conditions on its CoT traces varies substantially across models and tasks. 
These findings indicate that CoT steps do not always causally influence the final prediction, calling their faithfulness into question.

%% file: Figures/Figure2.tex
\begin{figure*}[t]
  \centering
  \includegraphics[width=1.0\textwidth]{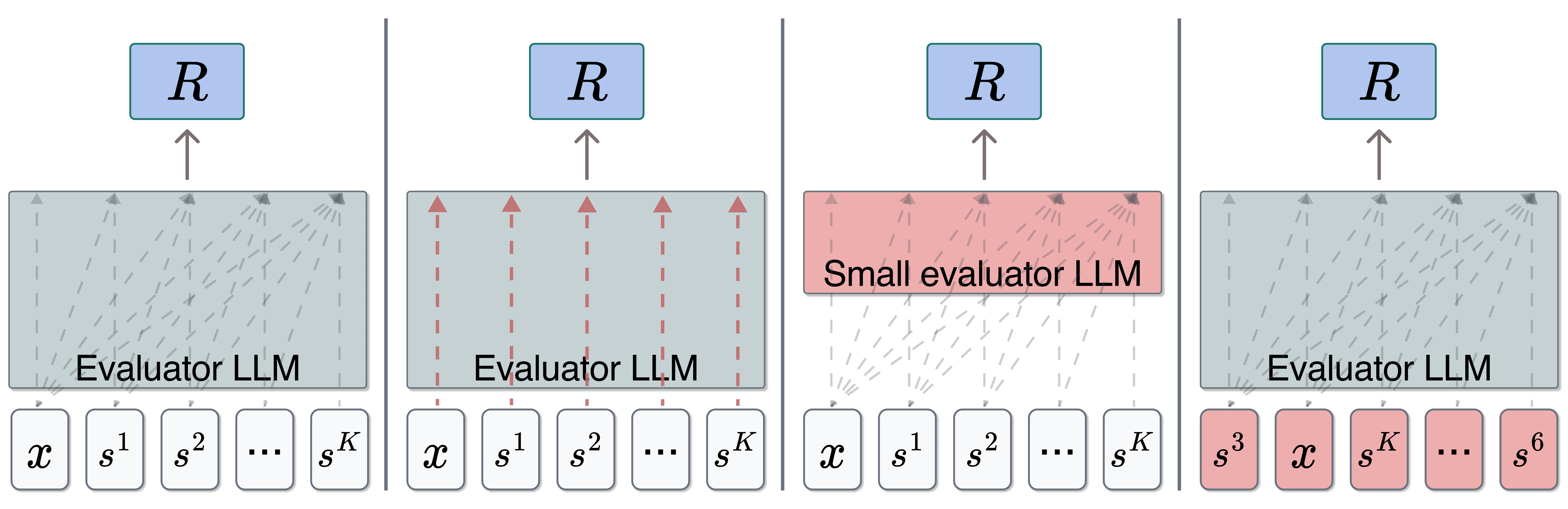}
  \caption{\textbf{Overall illustration of the proposed causality disruptions.} 
  From left to right, the diagrams depict: (i) the unaltered evaluation process, in which the full reasoning trace is evaluated autoregressively to calculate a probabilistic confidence metric; (ii) \textit{attention-level disruption}, where cross-step attention is masked and each reasoning step is evaluated independently; (iii) \textit{parameter-level disruption}, where a smaller evaluator model is used; and (iv) \textit{data-level disruption}, where the reasoning trace is modified (\textit{e.g.}, order shuffling) prior to evaluation.}
  \label{fig:fig2}
\end{figure*}

%% file: 3_methodology.tex
\section{Disruption of Probabilistic Confidence Metrics}

In this section, we introduce three approaches for disrupting probabilistic confidence metrics derived from the output token distribution of LLMs. 

\subsection{Preliminaries}

Let us denote an LLM as $\mathcal{M_\theta}$ and a given input prompt as $x$.
Given $x$, $\mathcal{M_\theta}$ generates a reasoning trace $T$ autoregressively.
Formally, let $T$ be a sequence of $K$ intermediate steps (\textit{i.e.}, $T=[\mathbf{s}^1,...,\mathbf{s}^K]$), where each step $\mathbf{s}^k$ is a sequence of ${L_k}$ tokens: $\mathbf{s}^k=[s^k_1, ...,s^k_{L_k}]$.
Each token $s^k_l$ is generated as
\begin{equation}
    s^k_l \sim p_\theta(\cdot|x,T_{<k},\mathbf{s}_{<l}^k),\label{eq:gen}
\end{equation}
where $T_{<k}$ denotes the previously generated reasoning steps and $\mathbf{s}_{<l}^k$ denotes the previously generated tokens within the current reasoning step.
The goal of Best-of-$N$ selection is to find the optimal reasoning trace $T^*$ among $N$ candidates $\{T_i\}_{i=1}^{N}$ sampled from $\mathcal{M}_{\theta}$, according to a scoring function $R$:
\begin{equation}
    T^* = \arg\max_{i=1,\dots,N} R(T_i).
\end{equation}
While $R$ can be implemented in a variety of ways, such as through external reward models or voting schemes, our work focuses on \textit{probabilistic confidence metrics}. 
For example, \citet{kang2025scalable} propose \textit{self-certainty} as a primary confidence metric for Best-of-$N$ selection:
\begin{equation*}
    R_{\texttt{SC}}=\frac{1}{nV} \sum_{k=1}^{K}\sum_{l=1}^{L_k}\sum_{j=1}^{V}\log\big(p_\theta(j|x,T_{<k},\mathbf{s}_{<l}^k)\big),
\end{equation*}
where $n=\sum_{k=1}^{K}L_k$. 

The generating LLM (Eq.~\ref{eq:gen}) and the evaluating LLM used to compute metrics are typically assumed to be the same model; however, this is not required. A reasoning trace $T_i$ may be sampled from a larger LLM, while the score $R$ is computed using a smaller model (see Section~\ref{sec:4.2} for the relevant experiments).

\subsection{Disruption of causality in reasoning traces}

Next, we introduce three forms of causality disruption applied to a multi-step reasoning trace $T$. 
We define \textit{causality disruption} as an intervention that weakens or removes the dependency structure between reasoning steps, while preserving surface-level logical fluency and semantic plausibility. 
Such disruptions are designed to reduce the degree to which later reasoning steps are conditionally dependent on earlier steps, without introducing grammatical or semantic artifacts.

Using these disruptions, we investigate the following research question: \textit{Are probabilistic confidence measures effective in selecting correct reasoning traces because they capture logical causality between reasoning steps?} 
To address this question, we design causality disruptions at three levels: \textbf{Attention, Parameter, and Data}.

\paragraph{Attention-level disruption.} 
\label{sec:3.2}

Probabilistic confidence metrics are derived from LLMs' token-level output distributions, which are calculated autoregressively. 
This induces confidence to be aggregated along the entire reasoning trace. 
As a result, earlier reasoning steps causally influence the probability scores of subsequent reasoning steps. 
Motivated by this observation, we introduce an \textit{attention-level causal disruption} by imposing a hard attention mask between steps within a reasoning trace when calculating the output token distribution and associated confidence metrics. 

Specifically, \textit{masked metrics} are computed by performing an independent forward pass for each reasoning step. 
We then obtain the final ‘‘masked’’ score by averaging step-level scores. 
Formally, the log-likelihood of a given input prompt $x$ is originally computed as:
\begin{equation*}
    R_{\tt LL}=\frac{1}{n}\sum\limits_{k=1}^{K}\sum\limits_{l=1}^{L_k} \log p_\theta(s_l^k|x,T_{<k},\mathbf{s}_{<l}^k),    
\end{equation*}
where the inter-step causal chain ($x \to \mathbf{s}^1 \to \mathbf{s}^2 \to \cdots \to \mathbf{s}^K$) is preserved during computation. 
To apply a disruption, we instead calculate:
\begin{equation}
    \widetilde{R}_{\tt LL}=\frac{1}{n}\sum\limits_{k=1}^{K}\sum\limits_{l=1}^{L_k} \log p_\theta(s_l^k|x,\mathbf{s}_{<l}^k),
\end{equation}
which breaks inter-step causality (\textit{i.e.}, $\mathbf{s}^{k-1} \not\to \mathbf{s}^k$) during metric calculation. 
Similarly, masked self-certainty and entropy are computed as:
\begin{equation}
    \widetilde{R}_{\texttt{SC}}=\frac{1}{nV} \sum_{k=1}^{K}\sum_{l=1}^{L_k}\sum_{j=1}^{V}\log\big(p_\theta(j|x,\mathbf{s}_{<l}^k)\big),
\end{equation}
\begin{equation}
    \widetilde{R}_{\tt ENT}=\frac{1}{n}\sum\limits_{k=1}^{K}\sum\limits_{l=1}^{L_k}H[p_\theta(\cdot|x,\mathbf{s}^k_{<l})],
\end{equation}
where $H[p(\cdot)]=-\sum_jp(j)\log p(j)$.
By removing conditioning on prior steps, tokens in step $\mathbf{s}^k$ cannot attend to tokens in $\mathbf{s}^{k-1}$, which is equivalent to forcing all cross-step attention weights to zero.
This disrupts the autoregressive causal chain by breaking the influence of earlier reasoning steps on later certainty estimates. 
The resulting metric therefore reflects the average local fluency of each individual step.

To further isolate local fluency signals, we also consider \textit{query-masked metrics} that additionally remove conditioning on the input query during metric calculation. 
For example, query-masked log-likelihood is computed as:
\begin{equation}
    \widehat{R}_{\tt LL}=\frac{1}{n}\sum\limits_{k=1}^{K}\sum\limits_{l=1}^{L_k} \log p_\theta(s_l^k|\mathbf{s}_{<l}^k).
\end{equation}
Query-masked self-certainty $\widehat{R}_{\tt SC}$ and entropy $\widehat{R}_{\tt ENT}$ are computed analogously.

\paragraph{Parameter-level disruption.}

Next, we introduce \textit{parameter-level disruption} by reducing the size and capacity of the evaluator model used to compute the output token distribution and associated probabilistic confidence metrics. 
Smaller LLMs have been shown to retain strong priors for common answer formats and local, sentence-level fluency, but are substantially limited in their ability to model long-range dependencies and complex causal structure present in multi-step reasoning traces \citep{wang2024comprehensive, ramezanali2025seqbench}. 
As a result, confidence estimates produced by such models are expected to emphasize local distributional regularities rather than global reasoning coherence. 
Specifically, we impose a parameter-level disruption by substituting the evaluator with \texttt{Qwen2.5-0.5B} -- a substantially smaller and less capable model relative to the generator models (\textit{e.g.}, \texttt{Qwen3-8B}) -- during metric computation.

\input{Figures/Tables/table1_masked.tex}

\paragraph{Data-level disruption.}
Finally, we introduce data-level disruptions by directly modifying the reasoning trace text prior to metric calculation. 
These modifications preserve sentence-level semantic meaning and linguistic fluency while distorting the inter-sentence dependency structure that encodes long-range reasoning. 
As a result, metrics are computed on locally coherent text that no longer reflects a causally coherent reasoning process. 
Specifically, we consider three forms of data-level causality disruptions: \textit{paraphrasing, shuffling, and truncation}, each targeting a different aspect of inter-step dependency.

\begin{itemize}[leftmargin=*, itemsep=0pt, topsep=0pt]  
    \item[$\circ$] \textit{Paraphrasing} disrupts causality by reducing syntactic dependencies between reasoning steps. 
    By independently rewriting each step in a reasoning trace, semantic content and logical claims are preserved, but the lexical and syntactic signals that link successive steps are altered. 
    This weakens cross-step dependencies without directly altering trace-level correctness or fluency.
    Each trace is segmented into reasoning steps and is individually passed to GPT-4o-mini \citep{achiam2023gpt} (detailed prompt in Appendix~\ref{app:paraphrasing-prompt}).
The following example illustrates how paraphrasing leads to inter-step disruptions:

\begin{tcolorbox}[
  title={Example of Paraphrasing},
  colback=white,
  colframe=black!40,
  boxrule=0.4pt
]
\textbf{Original:}  
(1) We isolate the variable \(x\).  
(2) We then square both sides to remove the square root.

\medskip

\textbf{Paraphrased:}  
(1) The variable \(x\) is separated from the remaining terms.  
(2) Squaring the equation eliminates the radical expression.
\end{tcolorbox}

\item[$\circ$] \textit{Shuffling} disrupts causality within a trace by perturbing the directionality of the reasoning process. 
Because the text of each reasoning step is unchanged, local step-level fluency is preserved.
However, the causal progression encoded by the ordering of steps is destroyed, eliminating sequential inter-step dependencies.
This is implemented by segmenting each trace into reasoning steps and then randomly shuffling their order. 

\begin{tcolorbox}[
  title={Example of Sentence-level Shuffling},
  colback=white,
  colframe=black!40,
  boxrule=0.4pt
]
\textbf{Original:}  
(1) Identify the equation.  
(2) Square both sides.  
(3) Solve for \(x\).

\medskip

\textbf{Shuffled:}  
(3) Solve for \(x\).  
(1) Identify the equation.  
(2) Square both sides.
\end{tcolorbox}

\item[$\circ$] \textit{Truncation} disrupts causality by removing the last few steps of the reasoning trace. 
The remaining text retains internal fluency and ordering, but because the causal chain is prematurely terminated, earlier reasoning is prevented from conditioning on later steps.
Each reasoning trace text is truncated to a set number of characters. 

\end{itemize}

%% file: Figures/Tables/table1_masked.tex
\begin{table*}[t]
    \centering
    \caption{Selection accuracy under original, masked, and query-masked (q-masked) metric evaluation.}
    \label{tab:tab1}
    \begin{adjustbox}{width=\textwidth}
    \begin{tabular}{lccccccccc}
        \toprule
        \multirow{2}{*}{\textbf{Benchmark}} 
        & \multicolumn{3}{c}{\textbf{Qwen}} 
        & \multicolumn{3}{c}{\textbf{Llama}} 
        & \multicolumn{3}{c}{\textbf{Phi}} \\
        \cmidrule(lr){2-4} \cmidrule(lr){5-7} \cmidrule(lr){8-10}
        & original & masked & q-masked & original & masked & q-masked & original & masked & q-masked \\
        \midrule
        MATH-500 & 95.60 & 95.40 & 95.60 & 46.20 & 39.40 & 36.40 & 49.60 & 46.20 & 43.80 \\
        GPQA-D   & 52.02 & 45.45 & 42.93 & 29.29 & 26.77 & 24.24 & 21.72 & 24.24 & 22.73 \\
        GSM8K    & 36.39 & 36.39 & 36.32 & 30.10 & 26.23 & 24.94 & 32.68 & 32.15 & 31.99 \\
        LogiQA   & 52.23 & 47.00 & 48.54 & 38.86 & 36.10 & 35.79 & 36.87 & 35.79 & 36.56 \\
        AR-LSAT  & 22.61 & 25.65 & 23.48 & 15.65 & 20.00 & 20.43 & 20.87 & 22.17 & 25.22 \\
        \midrule
        \textbf{Average} & 51.77 & 49.98 & 49.37 & 32.02 & 29.70 & 28.36 & 32.35 & 32.11 & 32.06 \\
        \bottomrule
    \end{tabular}
    \end{adjustbox}
\end{table*}

%% file: 4_results.tex
\section{Probabilistic Confidence Does Not Capture Causality in Reasoning}\label{sec:4}

If probability-based selection depended on inter-step reasoning coherence, disrupting the causal structure should substantially degrade selection accuracy. 
However, across extensive experiments, we find that \textbf{accuracy remains largely stable under our proposed causality disruptions.} 
Unless stated otherwise, all reported accuracies are averaged across outputs selected using self-certainty, log-likelihood, and entropy to reduce metric-specific variance. Full results are presented in Appendix~\ref{app:small_shuffled}.

\subsection{Experimental setup}

We conduct experiments using three representative open-source LLMs: (1) Qwen3-8B \citep{yang2025qwen3}, (2) Llama-3.1-8B-Instruct \citep{dubey2024llama}, and (3) Phi-3.5-mini-instruct \citep{phi3}. 
For brevity, we refer to these models as Qwen, Llama, and Phi for the remainder of the paper. 
We evaluate all models on open-ended (MATH-500 \citep{lightman2023let}, GSM8K \citep{cobbe2021gsm8k}) and multiple-choice reasoning benchmarks (GPQA-Diamond \citep{rein2024gpqa}, AR-LSAT \citep{zhong2021ar}, LogiQA \citep{liu2020logiqa}).\footnote{For brevity, we refer to GPQA-Diamond as GPQA-D.}

To generate CoT reasoning traces, we use zero-shot CoT prompting \citep{kojima2022large} to sample $N=10$ outputs per question with temperature $\tau=0.8$. 
Each output consists of a multi-step reasoning trace followed by a final answer. 
Reasoning steps are defined by segmenting the reasoning trace at punctuation boundaries \citep{bogdan2025thought}. 
We use normalized log-likelihood and entropy as representative sentence-level and distributional confidence measures, respectively. 
We also include self-certainty \citep{kang2025scalable}, a recently proposed state-of-the-art metric. 
Accuracy on a specific benchmark is computed as the fraction of questions for which the selected output is correct.

\subsection{Results of attention-level disruptions}

Table~\ref{tab:tab1} reports selection accuracy under progressively stronger attention-level causality disruptions. Across a variety of benchmarks and models, Best-of-$N$ selection accuracy remains largely unaffected by masking-based attention-level disruptions.
This stability is particularly evident for Qwen and Llama, with only modest changes (\(\leq 1.3\%\)) in accuracy throughout most benchmarks. For Qwen, masking even yields a small \textit{improvement} of 0.75\% in overall average accuracy, indicating that selection is largely insensitive to cross-step attention.
Llama shows a modest monotonic decrease in overall accuracy as masking severity increases (29.05\% $\to$ 28.69\% $\to$ 27.37\%). Even then, differences within individual benchmarks remain within 1--2\% in most cases.

Notably, Phi exhibits a consistently larger degradation, showing a 6.9\% decrease under masking and a 6.57\% decrease when the query is further masked. 
This pattern is consistent across multiple benchmarks, with particularly sharp declines on GPQA-Diamond and LogiQA, where accuracy reductions exceed 10\%. 
Nevertheless, even for Phi, degradation remains moderate relative to the severity of the intervention.
Despite fully blocking cross-step attention or re-conditioning each reasoning step independently on the input, selection accuracy does not collapse, and performance remains well above random selection baselines (e.g., Phi's average accuracy of 23.9\% on GPQA-Diamond). 

Overall, these results are consistent with the notion that selection using inherent probability-based metrics does not strongly rely on inter-sentence causal coherence.

\subsection{Results of parameter-level disruptions}\label{sec:4.2}

\input{Figures/Tables/table2_small.tex}

Table~\ref{tab:tab2} reports selection accuracy under parameter-level disruptions. 
Consistent with results from attention-level disruptions, selection accuracy remains largely stable.
For Qwen, performance changes by only 0.7\% on average, with individual accuracies remaining nearly identical across most benchmarks. Even on GPQA-Diamond, which exhibits the largest decline, accuracy drops by a modest 3.86\%.
Similar patterns hold for Llama and Phi. Llama's overall accuracy \textit{increases} by 0.3\%, and Phi's decreases only by 0.29\%. 

These results demonstrate that even when the evaluator model lacks fine-grained representational power or sophisticated reasoning ability, selection accuracy remains stable. 
This stability further supports the claim that probabilistic confidence metrics do not strongly rely on internal reasoning fidelity when ranking candidate outputs, and instead exploit coarse distributional signals such as token-level fluency, or answer format priors, which are preserved even in smaller evaluators.

\subsection{Results of data-level disruptions}
We now present results under data-level causality disruptions. 
For truncation and paraphrasing analyses, we narrow the experimental scope to Qwen evaluated on MATH-500, as both disruptions require reasoning traces that are sufficiently long and syntactically well-structured. 
Fixing the model and benchmark avoids confounding due to differences in output structure across tasks and model families, allowing for a clearer inspection of the effects of the proposed disruptions.

\paragraph{Sentence-level shuffling.}

\input{Figures/Tables/table3_shuffled.tex}

As shown in Table~\ref{tab:tab3}, shuffling the order of output sentences does not systematically degrade performance, and in some cases yields small improvements. 
At the benchmark level, shuffling induces both minor gains and losses. 
For example, Qwen improves on LogiQA (+2.03\%), Phi improves markedly on AR-LSAT (+3.46\%), while decreases are observed on GPQA-Diamond for Qwen and Llama. 
However, when aggregated, these effects offset each other, resulting in near-invariant averages. 
Averaged across benchmarks, Qwen and Llama exhibit virtually unchanged accuracy under shuffling (Qwen: 48.66\% $\to$ 48.40\%; Llama: 29.05\% $\to$ 29.01\%), while Phi shows a modest average increase (31.88\% $\to$ 32.96\%). 
Overall, the results indicate that preserving sentence-level linguistic fluency is largely sufficient for effective probability-based ranking, even when the causal order between reasoning steps is disrupted. 
This supports the claim that such metrics do not strongly rely on inter-sentence causal structure, instead reflecting coarse distributional properties that remain stable under sentence reordering.
\paragraph{Paraphrasing.}

\input{Figures/Figure3}

Figure~\ref{fig:fig3} shows that paraphrasing reasoning traces results in negligible change in Best-of-$N$ selection accuracy on Qwen-generated MATH-500 outputs.
Self-certainty and log-likelihood show only a 0.2\% decline in accuracy, and entropy accuracy remains unchanged after paraphrasing, suggesting that probabilistic confidence metrics are largely invariant to lexical and syntactic variations that preserve sentence-level meaning.

\paragraph{Truncation.}

\input{Figures/Figure4}

Figure~\ref{fig:fig4} illustrates the effect of progressively removing later reasoning steps from Qwen-generated outputs for MATH-500, while preserving the prefix of the trace. 
Here, truncation length refers to the maximum number of tokens retained from the beginning of each output during evaluation; tokens beyond this length are discarded before computing selection metrics.
By varying the truncation length, we progressively remove later portions of the reasoning trace while preserving the initial prefix, allowing us to assess how much of the output text is actually required for effective probability-based selection. 
In our setting, the average generated output length is approximately 22k characters, meaning that truncation lengths below this threshold remove substantial portions of the reasoning trace.

Across truncated lengths ranging from 1k to 30k tokens, selection accuracy remains largely stable, particularly for selection using self-certainty. 
Competitive selection performance is already achieved at relatively short lengths of 5k–10k tokens, indicating that access to a short prefix of the generated output is sufficient for effective ranking. 
Log-likelihood and entropy show slightly greater sensitivity, but \textit{decline} in accuracy as longer portions of the output are retained. 
Overall, the insensitivity of selection accuracy to truncation length suggests a low reliance on complete multi-step reasoning traces.

\subsection{Summary}

Across three different levels of disruption, we consistently observe that Best-of-$N$ selection using probabilistic confidence metrics is largely insensitive to disruptions of inter-step causal structure. 
In many cases, competitive performance is achieved using only locally fluent prefixes of the output, and certain causality disruptions actually improve performance. 
Taken together, these results reveal a \textbf{misalignment between probabilistic confidence metrics and reasoning quality.} 
The pronounced lack of sensitivity to inter-step causality disruptions suggests that these metrics are primarily exploiting coarse distributional signals, such as local fluency or answer format priors rather than faithfully capturing multi-step causal reasoning coherence. 
Consequently, current Best-of-$N$ practices likely overestimate the reliability of probabilistic confidence as a proxy for reasoning correctness.

%% file: Figures/Tables/table2_small.tex
\begin{table}[t]
    \centering
    \caption{Accuracy comparison for metric computation with the full model vs. a smaller model.}
    \label{tab:tab2}
    \begin{adjustbox}{width=\columnwidth, height=1.86cm}
    \begin{tabular}{lcccccc}
        \toprule
        & \multicolumn{2}{c}{\textbf{Qwen}} 
        & \multicolumn{2}{c}{\textbf{Llama}} 
        & \multicolumn{2}{c}{\textbf{Phi}} \\
        \cmidrule(lr){2-3} \cmidrule(lr){4-5} \cmidrule(lr){6-7}
        \textbf{Benchmark} & full & small & full & small & full & small \\
        \midrule
        MATH-500 & 95.60 & 95.00 & 46.20 & 36.80 & 49.60 & 44.00 \\
        GPQA-D   & 52.02 & 32.83 & 29.29 & 24.24 & 21.72 & 22.73 \\
        GSM8K    & 36.39 & 36.09 & 30.10 & 28.05 & 32.68 & 32.30 \\
        LogiQA   & 52.23 & 45.16 & 38.86 & 36.10 & 36.87 & 33.79 \\
        AR-LSAT  & 22.61 & 24.35 & 15.65 & 16.52 & 20.87 & 20.87 \\
        \midrule
        \textbf{Average} & \textbf{51.77} & 46.69 & \textbf{32.02} & 28.34 & \textbf{32.35} & 30.74 \\
        \bottomrule
    \end{tabular}
    \end{adjustbox}
\end{table}

%% file: Figures/Tables/table3_shuffled.tex
\begin{table}[t]
    \centering
    \caption{Accuracy comparison for metrics on ordered (ord.) vs. shuffled (shuf.) reasoning traces.}
    \label{tab:tab3}
    
    \begin{adjustbox}{width=\columnwidth}
    \begin{tabular}{lcccccc}
        \toprule 
        & \multicolumn{2}{c}{\textbf{Qwen}} 
        & \multicolumn{2}{c}{\textbf{Llama}} 
        & \multicolumn{2}{c}{\textbf{Phi}} \\
        \cmidrule(lr){2-3} \cmidrule(lr){4-5} \cmidrule(lr){6-7}
        \textbf{Benchmark} & ord. & shuf. & ord. & shuf. & ord. & shuf. \\
        \midrule
        MATH-500 & 95.60 & 95.60 & 46.20 & 41.40 & 49.60 & 48.80 \\
        GPQA-D   & 52.02 & 44.44 & 29.29 & 24.24 & 21.72 & 27.78 \\
        GSM8K    & 36.39 & 36.39 & 30.10 & 28.13 & 32.68 & 32.22 \\
        LogiQA   & 52.23 & 53.00 & 38.86 & 36.87 & 36.87 & 36.71 \\
        AR-LSAT  & 22.61 & 20.87 & 15.65 & 20.00 & 20.87 & 27.39 \\
        \midrule
        \textbf{Average} & \textbf{51.77} & 50.06 & \textbf{32.02} & 30.13 & 32.35 & \textbf{34.58} \\
        \bottomrule
    \end{tabular}
    \end{adjustbox}
\end{table}

%% file: Figures/Figure3.tex
\begin{figure}[t]
    \centering
    \includegraphics[width=\columnwidth, height=4.1cm]{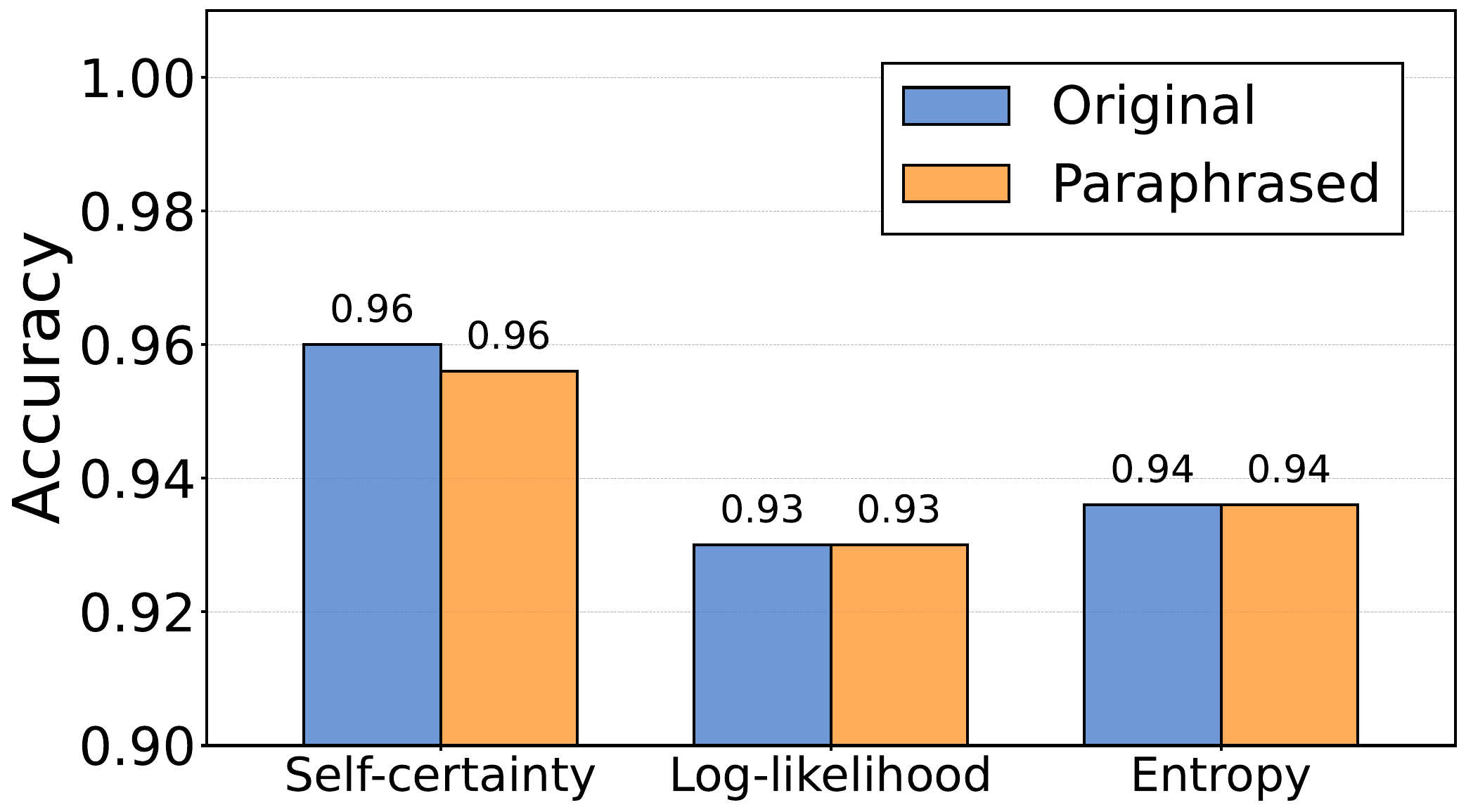}
    \caption{Paraphrased vs. original text selection accuracy on MATH-500 with Qwen. Paraphrasing leads to negligible performance drops.}
    \label{fig:fig3}
\end{figure}

%% file: Figures/Figure4.tex
\begin{figure}[t]
    \centering
    \includegraphics[width=1.0\columnwidth]{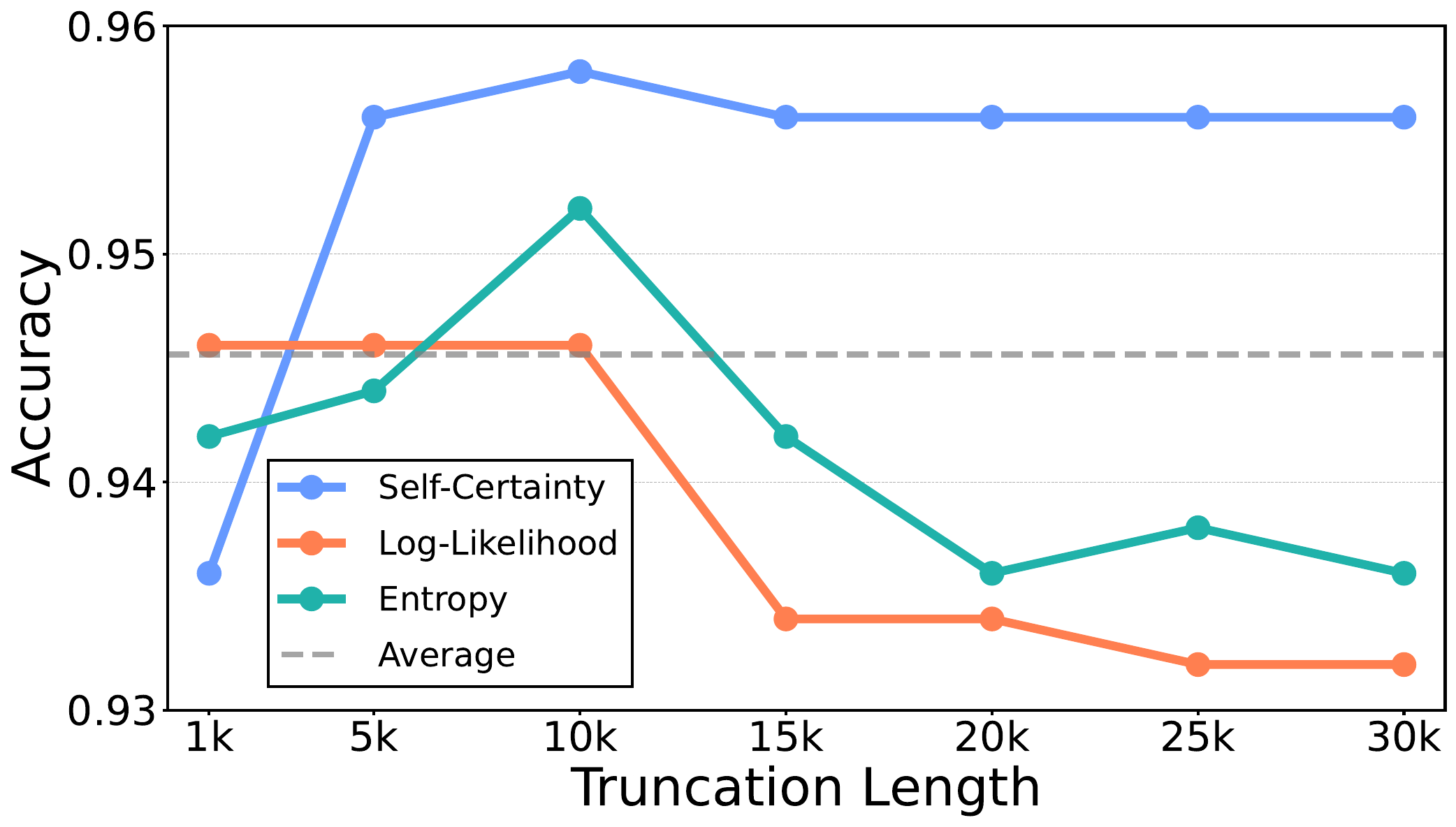}
    \caption{Selection accuracy as a function of truncation length on MATH-500 with Qwen. Heavy truncation does not substantially degrade accuracy and can even lead to improvements.}
    \label{fig:fig4}
\end{figure}

%% file: 5_newmetric.tex
\section{Contrastive Causality Metric}

Motivated by the observations in Section~\ref{sec:4}, we propose a new probabilistic confidence metric designed to isolate the causal contribution of inter-step dependencies.
Specifically, we introduce the \textit{contrastive causality metric}:
\begin{equation}
\label{eq:causal_score}
R_{\texttt{causal}}(y) = R(y) - \alpha \cdot \widehat{R}(y).
\end{equation}
Here, $R$ denotes the original probabilistic confidence metric (\textit{e.g.}, self-certainty), and $\widehat{R}$ denotes the masked metric computed by applying attention-level masking (Section~\ref{sec:3.2}). 
Both quantities are computed using the same evaluator LLM.
We treat $\alpha$ as a hyperparameter that controls the relative contribution of local fluency. 
We observe that results are stable across a wide range of values $\alpha \in [0.3, 0.7]$, and we report results using $\alpha = 0.5$ for simplicity. Detailed results can be found in Appendix~\ref{app:contrastive_full}.

The intuition behind this formulation is that standard confidence metrics capture both sentence-level fluency and inter-step reasoning consistency, whereas masked metrics are designed to explicitly remove inter-step causal effects. 
By taking the difference between these two quantities, we aim to decouple local fluency from the contribution of inter-step causal dependencies within probabilistic confidence metrics.

\input{Figures/Tables/table4_metric.tex}

Table~\ref{tab:tab4} compares selection accuracies between standard self-certainty and the proposed contrastive causality–based metric (implemented via query-masked self-certainty), averaged across models and evaluated in three conditions: no perturbation, evaluation on sentence-shuffled reasoning traces, and evaluation by a small LLM. 

When perturbations are not present during evaluation, contrastive causality shows an improvement in average selection accuracy. 
Selection using the metric attains an overall accuracy of 39.11\% compared to 38.72\% for self-certainty (+0.39\%). 
The improvement is driven primarily by GPQA-Diamond (+1.54\%) and LogiQA (+0.90\%), with a smaller but consistent edge on MATH-500 (+0.07\%). 
Overall, the aggregate trend supports the claim that the contrastive metric is more effective at identifying higher-quality reasoning traces than a purely probability-based self-certainty score.

Crucially, contrastive causality exhibits sensitivity to disruptions that weaken inter-sentence dependencies. 
Shuffling reasoning traces and evaluating confidence with a smaller evaluation model both lower selection accuracy (-0.17\% and -1.44\%, respectively). 
This effect is most pronounced on GPQA-Diamond, where contrastive accuracy decreases by 9.47\% from 35.87\% to 26.40\%, suggesting that the metric’s signal has a stronger reliance on the availability of coherent cross-sentence relations that are attenuated by the perturbation. 

%% file: Figures/Tables/table4_metric.tex
\begin{table}[t]
    \centering
    \caption{Accuracy comparison between Self-Certainty (SC) and Contrastive Causality-based selection (CC) using masked self-certainty, averaged across models.}
    \label{tab:tab4}
    
    \begin{adjustbox}{width=0.9\columnwidth}
    \begin{tabular}{llcc}
        \toprule
        \textbf{Benchmark} & \textbf{Perturbation} & \textbf{SC} & \textbf{CC} \\
        \midrule
        \multirow{3}{*}{MATH-500}
         & none       & \textbf{65.47} & 64.13 \\
         & shuffling  & \textbf{62.27} & 61.47 \\
         & small-eval & \textbf{62.40} & 61.40 \\
        \midrule
        \multirow{3}{*}{GPQA-D}
         & none       & \textbf{38.38} & 38.05 \\
         & shuffling  & \textbf{38.05} & 37.21 \\
         & small-eval & \textbf{37.71} & 36.36 \\
        \midrule
        \multirow{3}{*}{GSM8K}
         & none       & \textbf{90.27} & 89.89 \\
         & shuffling  & \textbf{87.92} & 87.54 \\
         & small-eval & \textbf{88.25} & 87.09 \\
        \midrule
        \multirow{3}{*}{LogiQA}
         & none       & \textbf{52.64} & 52.59 \\
         & shuffling  & 52.23 & \textbf{52.33} \\
         & small-eval & 51.31 & \textbf{52.28} \\
        \midrule
        \multirow{3}{*}{AR-LSAT}
         & none       & \textbf{46.09} & 44.93 \\
         & shuffling  & 45.80 & \textbf{46.09} \\
         & small-eval & 41.45 & \textbf{43.62} \\
        \bottomrule
    \end{tabular}
    \end{adjustbox}
\end{table}

%% file: 6_conclusion.tex
\section{Conclusion}

In this paper, we demonstrate an unexpected vulnerability of probabilistic confidence metrics: despite strong empirical effectiveness in Best-of-$N$ selection, these metrics remain largely insensitive to causality disruptions that weaken inter-step dependencies while preserving surface-level fluency. 
Motivated by this gap, we introduce a contrastive causality metric that explicitly isolates signals attributable to inter-step causal dependencies. 
This metric not only outperforms existing probability-based baselines, but maintains sensitivity to causality perturbations. 
Taken together, our results highlight a misalignment between probabilistic confidence and reasoning quality, and show that enforcing causal contrast can provide a principled path toward more faithful output selection.

\newpage

\section*{Broader Impact and Ethical Implications}

This work examines the reliability of probabilistic confidence metrics in selecting reasoning outputs from LLMs. 
By highlighting cases in which widely used metrics fail to capture the causal aspects of reasoning, our findings aim to improve the interpretability and reliability of LLMs. 
More reliable selection mechanisms may reduce the risk of deploying systems that appear confident while producing logically invalid or misleading reasoning, which is particularly important in high-stakes domains such as education, scientific analysis, and decision support.

At the same time, improved confidence-based selection methods could be misused to amplify the apparent credibility of incorrect or biased model outputs if applied without appropriate task-specific validation. 
Our results emphasize that probabilistic confidence should not be treated as a proxy for truth or correctness, and that evaluation metrics must be carefully stress-tested under controlled perturbations before deployment.

All experiments are conducted on publicly available datasets, and no personal or sensitive data are used. 
The evaluated models are employed strictly in an inference-only setting, without additional training or adaptation. 
We do not introduce new model capabilities, nor do we propose deployment in real-world decision-making systems.

%% file: 7_limitations.tex
\section*{Limitations} 

As our work focuses primarily on isolating the effect of perturbing multi-step causal dependencies between reasoning steps, we evaluate only on mathematical and scientific reasoning benchmarks that strongly emphasize these dependencies. 
Therefore, our experiments do not fully represent the diversity of reasoning phenomena exhibited by LLMs. 
In particular, we do not evaluate commonsense, multi-hop QA, or multi-modal reasoning benchmarks, due to the following reasons; 
we exclude commonsense reasoning benchmarks (\textit{e.g.}, Commonsense QA \citet{talmor2019commonsenseqa}, Winogrande \citet{sakaguchi2021winogrande}), because isolating reasoning quality in these settings is difficult due to confounding signals from implicit assumptions or surface-level plausibility. 
Also, we do not evaluate on multi-hop QA (\textit{e.g.}, HotpotQA \citet{yang2018hotpotqa}), or multimodal (\textit{e.g.}, CLEVR \citet{johnson2017clevr}, VCR \citet{qi2025vcr}) benchmarks where reasoning is distributed across multiple retrieved contexts, because selection metrics in such settings may depend more heavily on retrieval quality and context aggregation than on inter-sentence causal coherence. 
Addressing these settings would be a valuable direction for future work, to understand whether the observed insensitivity generalizes beyond text-based, multi-step reasoning tasks.

In addition, our study is primarily empirical and diagnostic in nature because it prioritizes interpretability and practical insight over theoretical completeness, and is intended to surface failure modes that may not be apparent from purely formal analyses.
As a result, our conclusions are based on observed behavior rather than theory.
While this approach limits the generality of our claims, it allows us to directly investigate the assumptions underlying widely used selection metrics, in realistic evaluation pipelines. 
Developing a theoretical framework that can explain these empirical observations, as well as methods that directly address the identified limitations of probability-based selection, remains an important direction for future work.

%% file: 8_appendix.tex
\appendix

\section{Experimental Details}

\subsection{Inference framework}
All reasoning traces are generated using \textsc{vLLM}. Models are loaded with their native chat templates and executed with a fixed GPU memory utilization of 0.9. The maximum generation length is capped by the model’s supported context length, and the temperature is set at $\tau = 0.8$. No top-$k$ or top-$p$ filtering is applied during generation.

\subsection{Dataset details and splits}
\label{app:dataset_splits}
All datasets used in this work are publicly available and were accessed under their original licenses. We use the test splits exclusively for evaluation and do not redistribute dataset contents.
We use test splits across all benchmarks. No training or development data are used for model fitting, calibration, or metric tuning.

\paragraph{Summary.}
Table~\ref{tab:dataset_statistics} summarizes the number of test examples used for each dataset.

\begin{table}[h]
\centering
\footnotesize
\caption{Dataset statistics and splits used in our experiments.}
\label{tab:dataset_statistics}
\setlength{\tabcolsep}{2pt}
\resizebox{\columnwidth}{!}{
\begin{tabular}{lcc}
\toprule
\textbf{Dataset} & \textbf{Task Type} & \textbf{\# Test Examples} \\
\midrule
MATH-500        & Math reasoning (open-ended)& 500 \\
GSM8K           & Math reasoning (open-ended)& 1{,}319 \\
GPQA-Diamond    & Science reasoning (MCQ)& 198 \\
LogiQA          & Logical reasoning (MCQ)     & 651 \\
AR-LSAT         & Analytical reasoning (MCQ)  & 230 \\
\bottomrule
\end{tabular}}
\end{table}

\subsection{Evaluation framework}
All probabilistic metrics are computed using an evaluator LLM in single full-sequence forward passes over the concatenated prompt and candidate output. Evaluator models are loaded with 4-bit NF4 quantization and evaluated with \texttt{use\_cache=False}. We compute normalized log-likelihood, top-$p$ entropy, and self-certainty from token-level log-probabilities, averaged over non-padding tokens. We additionally compute masked and query-masked metrics by re-evaluating individual steps either in isolation or conditioned on the prompt, and averaging scores across steps.

\subsection{GPU resources}
Generation and evaluation are performed on separate GPU hardware. All reasoning traces are generated on NVIDIA RTX 4090 GPUs, while all probabilistic confidence metrics and evaluator-model forward passes are computed on NVIDIA A6000 GPUs. 

\subsection{Detailed paraphrasing prompt}
We report the exact prompt used for sentence-level paraphrasing below.
\label{app:paraphrasing-prompt}
\begin{tcolorbox}[
  title={Paraphrasing Prompt},
  colback=white,
  colframe=black!30,
  boxrule=0.3pt,
  left=6pt,right=6pt,top=4pt,bottom=4pt
]

\textbf{Paraphrase the following reasoning sentence.}

\textbf{Rules:}
\begin{enumerate}
\item Preserve \textbf{all mathematical meaning} and symbols.
\item Keep logical relationships intact.
\item Make the wording \textbf{formal and clear}.
\item Change phrasing, syntax, and structure \textbf{as much as possible}.
\item Output \textbf{only one rewritten sentence}.
\end{enumerate}

\textbf{Sentence:}
\end{tcolorbox}

\section{Additional Results}
\subsection{$\alpha$ sensitivity of the contrastive causality metric}
\label{app:alpha_sensitivity}

We present an ablation study on the weighting parameter $\alpha$ used in the contrastive causality metric. Recall that the metric interpolates between standard self-certainty and attention-masked signal as defined in Eq.~\ref{eq:causal_score}. In particular, when $\alpha = 0$, the metric exactly reduces to the original self-certainty score, while larger values of $\alpha$ place increasing emphasis on contrastive causal consistency.

Table~\ref{tab:alpha_sensitivity} reports selection accuracy for each $\alpha$ for Qwen, Llama, and Phi. Performance remains stable for all three models across a wide range of $\alpha$ values ($0.1$--$0.9$). The maximum variation in accuracy is small relative to overall performance levels, indicating that the metric is not highly sensitive to precise tuning of $\alpha$.

\begin{table}[t]
\centering
\footnotesize
\caption{Accuracy as a function of the weighting parameter $\alpha$ for the contrastive causality metric. Results are averaged across benchmarks.}
\label{tab:alpha_sensitivity}
\setlength{\tabcolsep}{6pt}
\begin{tabular}{lccc}
\toprule
$\alpha$ & Qwen & Llama & Phi \\
\midrule
0.0 & .834 & .471 & .452 \\
0.1 & .830 & .472 & .451 \\
0.2 & .830 & .464 & .448 \\
0.3 & .830 & .465 & .447 \\
0.4 & .830 & .464 & .447 \\
0.5 & .830 & .459 & .448 \\
0.6 & .830 & .457 & .449 \\
0.7 & .830 & .457 & .449 \\
0.8 & .830 & .456 & .453 \\
0.9 & .830 & .450 & .455 \\
1.0 & .830 & .449 & .448 \\
\bottomrule
\end{tabular}
\end{table}

All accuracies reported in Table~\ref{tab:alpha_sensitivity} are computed using the \emph{query-masked} variant of the contrastive causality metric.

\label{app:small_shuffled}

\subsection{Full evaluation results for contrastive causality metrics}
\label{app:contrastive_full}
Table~\ref{tab:q_masked_metrics} reports accuracies for standard and query-masked variants of contrastive causality metrics across different probability metrics. \textbf{masked} denotes that the proposed metric is implemented using a masked metric, while the suffix \textbf{(q-)} indicates implementation under query masking.

\begin{table*}[t]
\centering
\footnotesize
\caption{Comparison of masked and query-masked variants of selection metrics. \textbf{certainty} denotes the contrastive causality metric using masked self-certainty.}
\label{tab:q_masked_metrics}
\setlength{\tabcolsep}{6pt}
\begin{tabular}{llcccccc}
\toprule
\textbf{Model} & \textbf{Benchmark} &
\multicolumn{2}{c}{\textbf{Self-certainty}} &
\multicolumn{2}{c}{\textbf{Log-likelihood}} &
\multicolumn{2}{c}{\textbf{Entropy}} \\
\cmidrule(lr){3-4} \cmidrule(lr){5-6} \cmidrule(lr){7-8}
 &  & masked & q-masked & masked & q-masked & masked & q-masked \\
\midrule
\multirow{5}{*}{Qwen}
 & MATH-500     & 0.9120 & 0.9618 & 0.9720 & 0.9579 & 0.9780 & 0.9579 \\
 & GPQA-Diamond & 0.5707 & 0.6061 & 0.5960 & 0.6010 & 0.5758 & 0.6010 \\
 & GSM8K        & 0.9500 & 0.9606 & 0.9644 & 0.9575 & 0.9644 & 0.9575 \\
 & LogiQA       & 0.6974 & 0.7358 & 0.7527 & 0.7296 & 0.7573 & 0.7158 \\
 & AR-LSAT      & 0.7304 & 0.8087 & 0.8522 & 0.7957 & 0.8565 & 0.8261 \\
\midrule
\multirow{5}{*}{Phi}
 & MATH-500     & 0.4100 & 0.4020 & 0.3980 & 0.3980 & 0.4260 & 0.3960 \\
 & GPQA-Diamond & 0.2879 & 0.3081 & 0.2727 & 0.2929 & 0.2778 & 0.2980 \\
 & GSM8K        & 0.8279 & 0.8006 & 0.8165 & 0.8105 & 0.8294 & 0.8082 \\
 & LogiQA       & 0.4086 & 0.4025 & 0.4178 & 0.4055 & 0.4025 & 0.4055 \\
 & AR-LSAT      & 0.2478 & 0.2478 & 0.2609 & 0.2565 & 0.2609 & 0.2609 \\
\midrule
\multirow{5}{*}{Llama}
 & MATH-500     & 0.4640 & 0.4140 & 0.4400 & 0.4000 & 0.4560 & 0.4120 \\
 & GPQA-Diamond & 0.3131 & 0.2424 & 0.2424 & 0.2374 & 0.2828 & 0.2323 \\
 & GSM8K        & 0.8635 & 0.8226 & 0.8294 & 0.7930 & 0.8650 & 0.8173 \\
 & LogiQA       & 0.4255 & 0.3641 & 0.4025 & 0.4086 & 0.4439 & 0.3948 \\
 & AR-LSAT      & 0.2043 & 0.2174 & 0.2652 & 0.2130 & 0.2304 & 0.2174 \\
\midrule
\textbf{Average} &  & 0.5542 & 0.5530 & 0.5655 & 0.5505 & 0.5738 & 0.5534 \\
\bottomrule
\end{tabular}
\end{table*}

\subsection{Full data- and parameter-level perturbation results}
For completeness, we report the full set of evaluation results for both data- and parameter-level perturbations across all benchmarks and models in Tables~\ref{tab:small_model_results_appendix} and~\ref{tab:shuffling_results_appendix}, respectively.

\begin{table*}[t]
\centering
\small
\caption{Selection accuracy under small-model evaluation. Results are reported for full- and small-model settings across probabilistic confidence metrics.}
\label{tab:small_model_results_appendix}
\setlength{\tabcolsep}{6pt}
\begin{tabular}{llcccccc}
\toprule
\textbf{Model} & \textbf{Benchmark}
& \multicolumn{2}{c}{\textbf{Self-certainty}}
& \multicolumn{2}{c}{\textbf{Log-likelihood}}
& \multicolumn{2}{c}{\textbf{Entropy}} \\
\cmidrule(lr){3-4} \cmidrule(lr){5-6} \cmidrule(lr){7-8}
& & \textbf{full} & \textbf{small}
& \textbf{full} & \textbf{small}
& \textbf{full} & \textbf{small} \\
\midrule

\multirow{5}{*}{Qwen}
& MATH-500     & 0.9760 & 0.9480 & 0.9580 & 0.9400 & 0.9560 & 0.9420 \\
& GPQA-Diamond & 0.6111 & 0.5859 & 0.5657 & 0.5657 & 0.5758 & 0.5657 \\
& GSM8K        & 0.9651 & 0.9583 & 0.9636 & 0.9560 & 0.9636 & 0.9545 \\
& LogiQA       & 0.7634 & 0.7235 & 0.7512 & 0.7296 & 0.7512 & 0.7343 \\
& AR-LSAT      & 0.8565 & 0.7522 & 0.8174 & 0.7696 & 0.8000 & 0.7826 \\
\midrule

\multirow{5}{*}{Llama}
& MATH-500     & 0.5220 & 0.4840 & 0.4560 & 0.4620 & 0.4660 & 0.4560 \\
& GPQA-Diamond & 0.2475 & 0.2323 & 0.2222 & 0.2323 & 0.2121 & 0.2071 \\
& GSM8K        & 0.8946 & 0.8605 & 0.8893 & 0.8696 & 0.8939 & 0.8605 \\
& LogiQA       & 0.4286 & 0.4224 & 0.4240 & 0.4301 & 0.4332 & 0.4163 \\
& AR-LSAT      & 0.2609 & 0.2348 & 0.2696 & 0.2522 & 0.2826 & 0.2522 \\
\midrule

\multirow{5}{*}{Phi}
& MATH-500     & 0.4660 & 0.4400 & 0.4900 & 0.4420 & 0.4520 & 0.4360 \\
& GPQA-Diamond & 0.2929 & 0.3131 & 0.2879 & 0.2929 & 0.2929 & 0.2929 \\
& GSM8K        & 0.8484 & 0.8287 & 0.8605 & 0.8309 & 0.8362 & 0.8324 \\
& LogiQA       & 0.3871 & 0.3932 & 0.3963 & 0.3840 & 0.3840 & 0.3917 \\
& AR-LSAT      & 0.2652 & 0.2565 & 0.2435 & 0.2304 & 0.2522 & 0.2565 \\
\bottomrule
\end{tabular}
\end{table*}

\begin{table*}[t]
\centering
\small
\caption{Selection accuracy under ordered and sentence-level shuffled reasoning traces across different probabilistic confidence metrics.}
\label{tab:shuffling_results_appendix}
\setlength{\tabcolsep}{6pt}
\begin{tabular}{llcccccc}
\toprule
\textbf{Model} & \textbf{Benchmark} 
& \multicolumn{2}{c}{\textbf{Self-certainty}}
& \multicolumn{2}{c}{\textbf{Log-likelihood}}
& \multicolumn{2}{c}{\textbf{Entropy}} \\
\cmidrule(lr){3-4} \cmidrule(lr){5-6} \cmidrule(lr){7-8}
& & \textbf{ordered} & \textbf{shuffled}
& \textbf{ordered} & \textbf{shuffled}
& \textbf{ordered} & \textbf{shuffled} \\
\midrule

\multirow{5}{*}{Qwen}
& MATH-500      & 0.9760 & 0.9820 & 0.9580 & 0.9720 & 0.9560 & 0.9060 \\
& GPQA-Diamond  & 0.6111 & 0.6212 & 0.5657 & 0.6111 & 0.5758 & 0.5505 \\
& GSM8K         & 0.9651 & 0.9644 & 0.9636 & 0.9568 & 0.9636 & 0.9575 \\
& LogiQA        & 0.7634 & 0.7542 & 0.7512 & 0.7419 & 0.7512 & 0.7327 \\
& AR-LSAT       & 0.8565 & 0.8348 & 0.8174 & 0.8435 & 0.8000 & 0.8000 \\
\midrule

\multirow{5}{*}{Llama}
& MATH-500      & 0.5220 & 0.4620 & 0.4560 & 0.4280 & 0.4660 & 0.4260 \\
& GPQA-Diamond  & 0.2475 & 0.2525 & 0.2222 & 0.2020 & 0.2121 & 0.2172 \\
& GSM8K         & 0.8946 & 0.8552 & 0.8893 & 0.8582 & 0.8939 & 0.8696 \\
& LogiQA        & 0.4286 & 0.4209 & 0.4240 & 0.4240 & 0.4332 & 0.4163 \\
& AR-LSAT       & 0.2609 & 0.2478 & 0.2696 & 0.2565 & 0.2826 & 0.2478 \\
\midrule

\multirow{5}{*}{Phi}
& MATH-500      & 0.4660 & 0.4240 & 0.4900 & 0.4260 & 0.4520 & 0.4380 \\
& GPQA-Diamond  & 0.2929 & 0.2677 & 0.2879 & 0.2626 & 0.2929 & 0.2879 \\
& GSM8K         & 0.8484 & 0.8180 & 0.8605 & 0.8143 & 0.8362 & 0.8241 \\
& LogiQA        & 0.3871 & 0.3917 & 0.3963 & 0.3978 & 0.3840 & 0.4025 \\
& AR-LSAT       & 0.2652 & 0.2913 & 0.2435 & 0.2696 & 0.2522 & 0.2478 \\
\bottomrule
\end{tabular}
\end{table*}

\section{Usage of AI Assistants}
In preparing this work, we utilized AI-based writing assistants to refine sentence structure, correct grammatical errors, and enhance readability. These tools were employed only for rephrasing and language improvements, ensuring that the technical content, methodology, and experimental findings remained entirely authored by the researchers. The use of AI assistance was limited to editorial enhancements without influencing the originality or scientific contributions of the paper.